\documentstyle[12pt,mkstyle]{article}

\newcommand{\Sektion}[1]
{\quad\\
\addtocounter{section}{1}

\noindent{\large\bf \thesection. #1}\\

}
\title{
Beliefs in Markov Trees - From Local Computations to Local Valuation
}
\shorttitle{Beliefs in Markov Trees - Towards Local Valuation}
\date{}
\newcommand{\Bem}[1]{}
\newcommand{\LitStelle}[1]{  

\vspace{-2.5mm}

\bibitem{#1}  }

\newcommand{\oantidot}{\overline{\odot}}
\newcommand{\V}{\mbox{\bf V}}
\newcommand{\AbbEins}
{
\begin{figure}
a) \begin{picture}(4,4.0)
\put(1,2){\circle*{0.2}} 
\put(0.7,2.3){$A$} 
\put(2,3){\circle*{0.2}} 
\put(2.3,3.3){$B$} 
\put(3,2){\circle*{0.2}} 
\put(3.3,2.3){$C$} 
\put(3,1){\circle*{0.2}} 
\put(3.3,0.7){$D$} 
\put(1,1){\circle*{0.2}} 
\put(0.7,0.7){$E$} 
\put(0.5,1.6){\line(1,0){3}}
\put(0.5,3.6){\line(1,0){3}}
\put(0.5,1.6){\line(0,1){2}}
\put(3.5,1.6){\line(0,1){2}}
\put(2.7,0.3){\line(1,0){1}}
\put(2.7,0.3){\line(0,1){2}}
\put(3.7,2.3){\line(-1,0){1}}
\put(3.7,2.3){\line(0,-1){2}}
\put(0.7,0.7){\line(1,0){2.6}}
\put(0.7,0.7){\line(0,1){0.6}}
\put(3.3,1.3){\line(-1,0){2.6}}
\put(3.3,1.3){\line(0,-1){0.6}}
\put(0.3,0.3){\line(1,0){1}}
\put(0.3,0.3){\line(0,1){2}}
\put(1.3,2.3){\line(-1,0){1}}
\put(1.3,2.3){\line(0,-1){2}}

\end{picture}
b) 
\begin{picture}(5,5.0)
\put(1,2){\circle*{0.2}} 
\put(0.7,2.3){$A$} 
\put(2,3){\circle*{0.2}} 
\put(2.3,3.3){$B$} 
\put(3,2){\circle*{0.2}} 
\put(3.3,2.3){$C$} 
\put(4.0,1){\circle*{0.2}} 
\put(4.3,0.7){$D$} 
\put(4,3){\circle*{0.2}} 
\put(4.3,2.7){$F$} 
\put(1,1){\circle*{0.2}} 
\put(0.7,0.7){$E$} 
\put(0.5,1.6){\line(1,0){3}}
\put(0.5,3.6){\line(1,0){3}}
\put(0.5,1.6){\line(0,1){2}}
\put(3.5,1.6){\line(0,1){2}}
\put(1.7,2.7){\line(1,0){2.6}}
\put(1.7,2.7){\line(0,1){0.6}}
\put(4.3,3.3){\line(-1,0){2.6}}
\put(4.3,3.3){\line(0,-1){0.6}}
\put(3.8,0.1){\line(0,1){3.8}}
\put(3.8,0.1){\line(1,0){0.8}}
\put(4.6,3.9){\line(0,-1){3.8}}
\put(4.6,3.9){\line(-1,0){0.8}}
\put(2.7,0.3){\line(1,0){1.5}}
\put(2.7,0.3){\line(0,1){2}}
\put(4.2,2.3){\line(-1,0){1.5}}
\put(4.2,2.3){\line(0,-1){2}}
\put(0.7,0.7){\line(1,0){3.6}}
\put(0.7,0.7){\line(0,1){0.6}}
\put(4.3,1.3){\line(-1,0){3.6}}
\put(4.3,1.3){\line(0,-1){0.6}}
\put(0.3,0.3){\line(1,0){1}}
\put(0.3,0.3){\line(0,1){2}}
\put(1.3,2.3){\line(-1,0){1}}
\put(1.3,2.3){\line(0,-1){2}}

\end{picture}
\caption{An example of hypergraphs a) with b) without compatible belief 
network}  \label{abbeins}
\end{figure}
} 
\newcommand{\AbbZwei}
{
\begin{figure}
a)
\begin{picture}(4,4.0)
\put(1,2){\circle*{0.2}} \put(0.7,2.3){$A$} 
\put(2,3){\circle*{0.2}} \put(2.3,3.3){$B$} 
\put(3,2){\circle*{0.2}} \put(3.3,2.3){$C$} 
\put(3,1){\circle*{0.2}} \put(3.3,0.7){$D$} 
\put(1,1){\circle*{0.2}} \put(0.7,0.7){$E$} 
\put(1,2){\vector(1,1){0.9}}
\put(3,2){\vector(-1,1){0.9}}
\put(3,2){\vector(0,-1){0.8}}
\put(3,1){\vector(-1,0){1.8}}
\put(1,1){\vector(0, 1){0.8}}

\end{picture}
b)
\begin{picture}(4,4.0)
\put(1,2){\circle*{0.2}} \put(0.7,2.3){$A$} 
\put(2,3){\circle*{0.2}} \put(2.3,3.3){$B$} 
\put(3,2){\circle*{0.2}} \put(3.3,2.3){$C$} 
\put(3,1){\circle*{0.2}} \put(3.3,0.7){$D$} 
\put(1,1){\circle*{0.2}} \put(0.7,0.7){$E$} 
\put(1,2){\vector(1,1){0.9}}
\put(3,2){\vector(-1,1){0.9}}
\put(3,1){\vector(0, 1){0.8}}
\put(3,1){\vector(-1,0){1.8}}
\put(1,1){\vector(0, 1){0.8}}

\end{picture}
c)
\begin{picture}(4,4.0)
\put(1,2){\circle*{0.2}} \put(0.7,2.3){$A$} 
\put(2,3){\circle*{0.2}} \put(2.3,3.3){$B$} 
\put(3,2){\circle*{0.2}} \put(3.3,2.3){$C$} 
\put(3,1){\circle*{0.2}} \put(3.3,0.7){$D$} 
\put(1,1){\circle*{0.2}} \put(0.7,0.7){$E$} 
\put(1,2){\vector(1,1){0.9}}
\put(3,2){\vector(-1,1){0.9}}
\put(3,1){\vector(0, 1){0.8}}
\put(1,1){\vector( 1,0){1.8}}
\put(1,1){\vector(0, 1){0.8}}

\end{picture}
d)
\begin{picture}(4,4.0)
\put(1,2){\circle*{0.2}} \put(0.7,2.3){$A$} 
\put(2,3){\circle*{0.2}} \put(2.3,3.3){$B$} 
\put(3,2){\circle*{0.2}} \put(3.3,2.3){$C$} 
\put(3,1){\circle*{0.2}} \put(3.3,0.7){$D$} 
\put(1,1){\circle*{0.2}} \put(0.7,0.7){$E$} 
\put(1,2){\vector(1,1){0.9}}
\put(3,2){\vector(-1,1){0.9}}
\put(3,1){\vector(0, 1){0.8}}
\put(1,1){\vector( 1,0){1.8}}
\put(1,2){\vector(0,-1){0.8}}

\end{picture}
\caption{An example of belief networks corresponding to a hypergraph from 
Fig.\ref{abbeins} a)}  \label{abbzwei}
\end{figure}
} 
\newcommand{\AbbVierPic}[3]
{\begin{picture}(9,6.0)
\put(1,2){\circle*{0.2}} 
\put(0.7,1.7){$Y_1$} 
\put(3,2){\circle*{0.2}} 
\put(2.7,1.7){$Y_2$} 
\put(5,2){\circle*{0.2}} 
\put(4.7,1.7){$Y_3$} 
\put(2,4){\circle*{0.2}} 
\put(1.7,4.3){$X_1$} 
\put(4,4){\circle*{0.2}} 
\put(4.3,4.3){$X_2$} 
\put(0.7,1.7){\line(0,1){3.1}}
\put(0.7,1.7){\line(1,0){4.7}}
\put(5.4,4.8){\line(0,-1){3.1}}
\put(5.4,4.8){\line(-1,0){4.7}}
                   \put(0.8,3.2){twig: #1}
\put(0.1,0.1){\line(0,1){2.8}}
\put(0.1,0.1){\line(1,0){5.8}}
\put(5.9,2.9){\line(0,-1){2.8}}
\put(5.9,2.9){\line(-1,0){5.8}}
                  \put(0.3,0.3){branch: #2}
\put(3.5,1.0){\circle{0.2}} 
\put(5.5,1.3){\circle{0.2}} 
\put(4.0,0.7){\line(0,1){1.6}}
\put(4.0,0.7){\line(1,0){2.8}}
\put(6.8,2.3){\line(0,-1){1.6}}
\put(6.8,2.3){\line(-1,0){2.8}}
         \put(4.8,2.5){other hyperedge: #3}
\put(6.0,1.0){\circle{0.2}} 
\put(6.5,2.0){\circle{0.2}} 
\end{picture}

}
\newcommand{\AbbVier}
{
\begin{figure}
a) \AbbVierPic{$BEL_t$}{$BEL_b$}{$BEL_o$}

b) \AbbVierPic{$BEL_t$
}{$(BEL_b\oantidot BEL_b ^{\downarrow t \cap b})\odot BEL_b ^{\downarrow t 
\cap b}$ 
}{$(BEL_o\oantidot BEL_o ^{\downarrow t \cap b \cap o})\odot BEL_o 
^{\downarrow t \cap b \cap o}$ 
}

c) \AbbVierPic{$BEL'_t=BEL_t\odot BEL_b ^{\downarrow t \cap b}\odot BEL_o 
^{\downarrow t \cap b \cap o} $
}{$BEL_b\oantidot BEL_b ^{\downarrow t \cap b}$ 
}{$BEL_o\oantidot BEL_o ^{\downarrow t \cap b \cap o}$ 
} 
\end{figure}

\begin{figure}
d) \AbbVierPic{${BEL'}_t ^{| t \cap b}\odot {BEL'}_t
^{\downarrow t \cap b} $
}{$BEL_b\oantidot BEL_b ^{\downarrow t \cap b}$ 
}{$BEL_o\oantidot BEL_o ^{\downarrow t \cap b \cap o}$ 
}

e) \AbbVierPic{${BEL'}_t ^{| t \cap b}$
}{$(BEL_b\oantidot BEL_b ^{\downarrow t \cap b})\odot {BEL'}_t
^{\downarrow t \cap b} $ 
}{$BEL_o\oantidot BEL_o ^{\downarrow t \cap b \cap o}$ 
} 

\caption{An example of valuation transformation}  
\label{abbvier} 
\end{figure}
} 

\newcommand{\AbbDrei}
{
\begin{figure}
a) \begin{picture}(6,6.0)
\put(1,2){\circle*{0.2}} 
\put(0.7,1.7){$Y_1$} 
\put(3,2){\circle*{0.2}} 
\put(2.7,1.7){$Y_2$} 
\put(5,2){\circle*{0.2}} 
\put(4.7,1.7){$Y_3$} 
\put(2,4){\circle*{0.2}} 
\put(1.7,4.3){$X_1$} 
\put(4,4){\circle*{0.2}} 
\put(4.3,4.3){$X_2$} 
\put(0.7,1.7){\line(0,1){3.1}}
\put(0.7,1.7){\line(1,0){4.7}}
\put(5.4,4.8){\line(0,-1){3.1}}
\put(5.4,4.8){\line(-1,0){4.7}}
\put(0.8,3.2){twig}
\put(0.1,0.1){\line(0,1){2.8}}
\put(0.1,0.1){\line(1,0){5.8}}
\put(5.9,2.9){\line(0,-1){2.8}}
\put(5.9,2.9){\line(-1,0){5.8}}
\put(0.3,0.3){branch}
\put(3.5,1.0){\circle{0.2}} 
\put(5.5,1.3){\circle{0.2}} 
\end{picture}
b) 
\begin{picture}(6,6.0)
\put(1,2){\circle*{0.2}} 
\put(0.7,1.7){$Y_1$} 
\put(3,2){\circle*{0.2}} 
\put(2.7,1.7){$Y_2$} 
\put(5,2){\circle*{0.2}} 
\put(4.7,1.7){$Y_3$} 
\put(2,4){\circle*{0.2}} 
\put(1.7,4.3){$X_1$} 
\put(4,4){\circle*{0.2}} 
\put(4.3,4.3){$X_2$} 
\put(2,4){\vector(1,0){1.8}} 
\put(1,2){\vector( 1,2){0.9}} 
\put(3,2){\vector(-1,2){0.9}} 
\put(5,2){\vector(-3,2){2.8}} 
\put(1,2){\vector( 3,2){2.8}} 
\put(3,2){\vector( 1,2){0.9}} 
\put(5,2){\vector(-1,2){0.9}} 
\end{picture}
\caption{
An example of a) a twig in hypertree and b) its fragment of belief 
network} \label{abbdrei} 
\end{figure}
} 

\begin{document}
\begin{center}
\large
\bf BELIEFS IN MARKOV TREES - FROM LOCAL COMPUTATIONS TO LOCAL VALUATION\\
\quad\\
\quad\\
MIECZYS{\L}AW A. K{\L}OPOTEK\\
\it Institute of Computer Science, Polish Academy of Sciences\\
\it PL 01-237 Warsaw, 21 Ordona St., Poland\\
\end{center}
%

\pagestyle{empty}

\begin{abstract}
{\small 
\baselineskip=12pt
 This paper is devoted to expressiveness of hypergraphs for which
uncertainty propagation by local computations via Shenoy/Shafer
method applies. It is demonstrated that for this propagation method 
for a given joint belief distribution 
no valuation of hyperedges of a hypergraph may provide with 
simpler hypergraph structure than 
valuation of hyperedges by conditional distributions. This has vital      
implication that methods recovering belief networks from data have no better 
alternative for finding the simplest hypergraph structure for belief 
propagation. A method for recovery tree-structured belief networks has 
been developed and specialized for Dempster-Shafer belief functions.
}
\end{abstract}

\baselineskip=13.7pt

\Sektion{INTRODUCTION}
Shenoy and Shafer \cite{Shenoy:90} presented an axiomatic system and a method 
for calculation of 
conditional beliefs by local computations. The method enables to reduce 
space requirements for representation of multivariate joint belief 
distributions.  Shenoy and Shafer 
 have shown that their framework is suitable both for bayesian 
and Dempster-Shafer belief distributions.
\Bem{
The major idea is the following: an explicit representation of a 
(discrete) joint belief distribution may be space-consuming, and if the number
of variables grows, it may easily grow impossible to be represented
even in large database systems. On the other hand only relatively few direct
dependencies may exist between one single variable and the remaining ones.
Hence it may be possible to factorize the whole distribution into a series 
of factors in smaller numbers of variables each where combination of factors
yields the same distribution, while the space required for representation
of all the factors together is drastically smaller than the one for the joint 
distribution as such.

Now given the factorization, we are keen on computing conditionals of some
few variables on some observations. We can win something from the 
factorization
if we were capable to calculate these conditionals without necessity
of restoring the joint distribution (and hence without all the space 
burden).

 Shenoy and Shafer \cite{Shenoy:90} demonstrated that this is actually 
possible 
if the factorization meets their axiomatic system. }
They generalize in their paper similar works of other authors who
concentrated on uncertainty propagation in bayesian networks
 \cite{Cannings:78}, \cite{Pearl:86b}, \cite{Lauritzen:88}, 
and for DS belief functions \cite{Shafer:87}, 
\cite{Hsia:89}.

One important innovation of Shenoy and Shafer \cite{Shenoy:90}
was to separate the notion of factorization from 
the notion of conditionality. Other authors insisted previously
 \cite{Pearl:86b}, \cite{Lauritzen:88} that factors in bayesian networks be 
conditional distributions.

The goal of this paper is to investigate to what extent this innovation
is really significant and what impact it may have on development
of algorithms for recovery (identification) of belief networks 
(factorization of belief distribution) from data.
\Bem{
The rest of this paper is organized as follows. 
In section 2 we recall the Shenoy/Shafer axiomatic framework.
In section 3 we consider the relationship between hypergraphs and
belief networks.
Section 4, the essential part of this paper, investigates the relationship
between hypertrees and belief networks.
Section 5 is devoted to a special type of belief network: the tree.
Conclusions are contained in section 6.
}
\Sektion{SHENOY/SHAFER \Bem{AXIOMATIC} FRAMEWORK}
We recall below some definitions from \cite{Shenoy:90}:

{\em Hypergraphs}: A nonempty set H of nonempty subsets of a finite set S be 
called 
a hypergraph on S. The elements of H be called hyperedges. Elements of S be 
called vertices. H and H' be both hypergraphs on S, then we call a 
hypergraph H' a {\em reduced hypergraph} of the 
hypergraph H, iff for every $h'\in H'$ also  $h'\in H$ holds, and for 
every  $h \in H$ there exists such a $h' \in H'$ that $h \subseteq h'$.
A hypergraph H {\em covers} a hypergraph H' iff for every $h'\in H'$ there 
exists such a $h\in H$ that $h'\subseteq h$.

{\em Hypertrees}: t and b be distinct hyperedges in a hypergraph H, $t \cap 
b\neq 
\emptyset$, and b contains every vertex of t that is contained in a hyperedge 
of H other than t; if $X\in t$ and $X\in h$, where $h\in H$ and $h\neq t$, 
then $X\in b$. Then we call t a twig of H, and we call b a branch for t. A 
twig may have more than one branch. 
We call a hypergraph a hypertree if there is an ordering of its hyperedges, 
say $h_1,h_2,...,h_n$ such that $h_k$ is a twig in the hypergraph 
$\{h_1,_h2,...,h_k\}$ whenever $2 \leq k \leq n$. We call  any  such 
ordering of 
hyperedges a hypertree construction sequence for the hypertree. The first 
hyperedge in the hypertree construction sequence be called the root of the 
hypertree construction sequence. 

{\em Factorization}: Suppose A is a valuation on a finite set of variables \V,
and suppose HV is a hypergraph on \V. If A is equal to the combination of 
valuations of all hyperedges h  of HV then we say that A factorizes on HV.

{\em Conditioning}: Suppose $BEL$ is a belief distribution,  and $BEL_E$ is 
an indicator potential   capturing the 
evidence E. Then conditional belief function conditioned on E, $BEL(.|E)$, is 
defined as  $BEL(.|E)=BEL \odot BEL_E$ (see \cite{Shenoy:90} p.191 for 
probabilistic case, $Bel_E$ may be a simple support function in DS case),
The axiom A3 states that to compute $(G \odot H) ^{\downarrow g}$ it is not 
necessary to compute $G \odot H$ first.

Shenoy and Shafer consider it unimportant whether or not the factorization 
should refer to conditional probabilities in case of probabilistic belief 
networks. 
But for expert system 
inference engine it is of primary importance how contents of 
knowledge base should be understood by a user
as it should at least  justify its conclusions 
by elements of knowledge base. So if a 
belief network (or a 
hypergraph)
 is to be used as  knowledge base, as much elements as possible 
have to refer to experience of the user.

In our opinion, the major reason for this remark of Shenoy and Shafer is that 
in fact the Dempster-Shafer belief function cannot be decomposed in terms of 
any conditional belief function as defined in the literature 
\cite{Klopotek:93p4}. But an intriguing question remains 
whether replacement of conditional belief function with a any belief 
function in the factorization extends essentially the class of such 
factorizations.
\Sektion{HYPERGRAPHS AND BELIEF NETWORKS}
The axiomatization system of Shenoy/Shafer refers to the notion of
factorization along a hypergraph.
On the other hand other authors insisted on a decomposition 
into a belief network. We investigate below implications
of this disagreement
%
%
\begin{df}
We define               a mapping $\oantidot: VV \times VV 
\rightarrow VV$ called decombination such that: 
if $BEL_{12}=BEL_1 \oantidot BEL_2$ then $BEL_1=BEL_2 \odot BEL_{12}$.
 \end{df}
In case of probabilities, decombination means memberwise division: 
$Pr_{12}(A)=Pr_1(A)/Pr_2(A)$. In case of DS pseudo-belief functions it means 
the operator $\ominus$ yielding a DS pseudo-belief function such that: 
whenever $Bel_{12}=Bel_1 \ominus Bel_2$ 
then $Q_{12}(A)=c \cdot Q_1/Q_2$. Both for probabilities and for DS belief 
functions decombination may be not uniquely determined. Moreover, for DS 
belief functions not always a decombined DS belief function will exist. Hence 
we extend the domain to DS pseudo-belief functions which is closed under this 
operator. We claim here without a proof (which is simple) that DS 
pseudo-belief 
functions fit the axiomatic framework of Shenoy/Shafer. Moreover, we claim 
that if an (ordinary) DS  belief  function  is  represented  by  a 
factorization in 
DS pseudo-belief functions, then any propagation of uncertainty yields the 
very 
same results as when it would have been factorized into ordinary DS belief 
functions. 
\begin{df}
By mk-conditioning $|$ of a belief function $BEL$ on a set of variables $h$ 
we understand the transformation: $BEL ^{|h}= BEL \oantidot BEL ^{\downarrow 
h}$. 
\end{df}
Notably, mk-conditioning means in case of probability functions proper 
conditioning. In case of DS pseudo-belief functions the operator $|$ has 
meaning entirely different from traditionally used notion of conditionality 
(compare 
\cite{Klopotek:93p4}) - mk-conditioning is a technical term used exclusively 
for valuation of nodes in belief networks. Notice: some other authors 
e.g. \cite{Cano:93} recognized also the necessity of introduction of two 
different notions in the context of the Shenoy/Shafer axiomatic framework 
(compare a priori and a posteriori conditionals in \cite{Cano:93}). 
\cite{Cano:93} introduces 3 additional axioms governing the 'a priori' 
conditionality to enable propagation with them.  
Our 
mk-conditionality is bound only to the assumption of executability of the 
$\oantidot$ operation and does not assume any further properties of it. 
We will discuss the consequences of this difference elsewhere.
Let 
us define now the general notion of belief networks (generalizing
definition of belief network from \cite{Geiger:90} - bayesian networks,
and \cite{Klopotek:93d} - DS networks). :
\AbbDrei
\begin{df}
 A 
belief 
 network is a pair (D,BEL) where D is a dag (directed acyclic graph)
and BEL  is a belief 
distribution called the {\em underlying distribution}. Each node i in D 
corresponds to a variable $X_i$  in BEL, a set of nodes I corresponds to a 
set of variables $X_I$ and $x_i, x_I$
 denote values drawn from the domain of $X_i$ 
 and from the (cross product) domain of $X_I$ respectively. Each node in the 
network  is regarded as a storage cell for any  distribution 
$BEL ^{\downarrow \{X_i\} \cup X_{\pi (i)} |  X_{\pi (i)} }$
 where $X_{\pi (i)}$ is a set of nodes corresponding to 
the 
parent nodes $\pi(i)$ of i.  The underlying distribution represented by a 
 belief network is computed via:
$$BEL  = \bigodot_{i=1}^{n}BEL ^{\downarrow \{X_i\} \cup X_{\pi (i)} |  
X_{\pi (i)} } $$
\end{df}
Please notice the local character of valuation of a node:
to valuate the node $i$ corresponding to variable $X_i$ only 
the marginal $BEL ^{\downarrow \{X_i\} \cup X_{\pi (i)}}$ needs to be known 
(e.g. from data) and not the entire belief distribution.

There exists a straight forward transformation of a belief network structure
into a hypergraph, and hence of 
a belief network into a hypergraph:
for every node i of the underlying dag define a hyperedge as the set
$\{X_i\} \cup X_{\pi(i)}$; then the valuation of this hyperedge define as
$BEL ^{\downarrow \{X_i\} \cup X_{\pi(i)} | X_{\pi(i)}}$. We say that the 
hypergraph obtained in this way is {\em induced} by the belief network.
 \AbbEins

Let us consider now the inverse operation: transformation of a valuated
hypergraph into a belief network.
As the first  stage we consider structures of a hypergraph and of a
belief network (the underlying dag). we say that a belief network is 
{\em compatible} with a hypergraph 
iff the reduced set of hyperedges induced by    the belief network is 
identical with the reduced hypergraph. 

\begin{Bsp} 
Let us consider the following hypergraph  (see Fig.\ref{abbeins}.a)):
\{\{A,B,C\}, \{C,D\}, \{D,E\}, \{A, E\}\}.
the following belief network structures are compatible with this hypergraph:
\{$A,C\rightarrow B$, $C\rightarrow D$, $D\rightarrow E$, $E\rightarrow A$\}
 (see Fig.\ref{abbzwei}.a)),
\{$A,C\rightarrow B$, $D\rightarrow C$, $D\rightarrow E$, $E\rightarrow A$\},
 (see Fig.\ref{abbzwei}.b)),
\{$A,C\rightarrow B$, $D\rightarrow C$, $E\rightarrow D$, $E\rightarrow A$\},
 (see Fig.\ref{abbzwei}.c)),
\{$A,C\rightarrow B$, $D\rightarrow C$, $E\rightarrow D$, $A\rightarrow E$\}.
 (see Fig.\ref{abbzwei}.d)),
\end{Bsp}

 \AbbZwei

\begin{Bsp}
Let us consider the following hypergraph  (see Fig.\ref{abbeins}.b)):
\{\{A,B,C\}, \{C,D\}, \{D,E\}, \{A, E\}, \{B,F\}, \{F,D\}\}.
No belief network structure is compatible with it.
\end{Bsp}
The missing compatibility is connected with the fact that
a hypergraph may represent a cyclic graph. 
Even if a compatible belief network has been found we may have troubles with 
 valuations. In Example 1 an unfriendly valuation of hyperedge 
\{A,C,B\} may require an edge AC in a belief network representing the same 
distribution, but  it will  make 
the 
hypergraph incompatible (as e.g. hyperedge \{A,C,E\} would be induced). This 
may be demonstrated as follows:
\begin{df}  
If $X_J,X_K,X_L$ are three disjoint sets of variables of a distribution BEL, 
then $X_J,X_K$ are said to be conditionally independent given $X_L$ (denoted 
\linebreak 
$I(X_J,X_K |X_L)_{BEL}$ iff 
 $$BEL ^{\downarrow X_J \cup X_K \cup X_L |  X_L} 
 \odot  BEL ^{\downarrow   X_L } =
 BEL ^{\downarrow X_J  \cup X_L |  X_L} \odot 
 BEL ^{\downarrow X_K \cup X_L |  X_L} 
 \odot  BEL ^{\downarrow   X_L } $$
%
$I(X_J,X_K |X_L)_{BEL}$ is called a {\em 
conditional independence statement}
\end{df}
Let $I(J,K|L)_D$ denote d-separation in a graph \cite{Geiger:90}.:
\begin{th} \label{IDIBEL}
Let $BEL_D=\{BEL|$(D,BEL) be a belief network\}. Then:\\
$I(J,K|L)_D$ iff $I(X_J,X_K |X_L)_{BEL}$ for all $BEL \in BEL_D$.
\end{th}
Proof of this theorem may be constructed analogously to the one for DS 
belief networks in \cite{Klopotek:93d}. 
Now we see in the above example that nodes D and E d-separate nodes A and C. 
 Hence within any belief network based on one of the three dags mentioned A 
will 
be conditionally independent from C given D and E. But one can easily check 
that with general type of hypergraph valuation nodes A and C may be rendered 
dependent. 
 The sad result of this section is, that really 
\begin{th}
Hypergraphs considered  by
 Shenoy/Shafer \cite{Shenoy:90} 
may for a given joint belief distribution have simpler structure than
(be properly covered by)
 the closest hypergraph induced by a 
belief network.
\end{th}
%
%
%
\Sektion{HYPERTREES AND BELIEF NETWORKS}
Notably, though the axiomatic system of Shenoy/Shafer refers to hypergraph
factorization of a joint belief distribution, the actual propagation is run 
on a hypertree (or more precisely, on one construction sequence of a 
hypertree, that is on Markov tree) covering that hypergraph. 
Covering a 
hypergraph with a hypertree is a trivial task, yet finding the optimal one 
(with 
as small number of variables in each hyperedge of the hypertree as possible) 
may be very difficult \cite{Shenoy:90}.

 Let us look closer at the outcome of the process of covering with a reduced
 hypertree factorization, or more precisely, at the relationship of a 
hypertree construction 
sequence and a  belief network constructed out of it in the following way:
If $h_k$ is a twig in the sequence $\{h_1,...,h_k\}$ and $h_{i_k}$ its branch 
with $i_k<k$, then let us span the following directed edges in a belief 
network: First make a complete directed acyclic graph out of nodes 
$h_k-h_{i_k}$. Then add edges $Y_l \rightarrow X_j$ for every $Y_l \in 
h_k \cap h_{i_k}$ and every $X_j \in h_k-h_{i_k}$.  (see Fig.\ref{abbdrei}).
Repeat this for every k=2,..,n. 
nodes contained  in $h_1$).

         For k=1 proceed as if $h_1$ were a twig with an
empty set as a  branch for it. 
 \AbbVier
 It is easily checked that 
the hypergraph induced by a belief network structure obtained in this way is 
in fact a hypertree (if reduced, then exactly the original reduced 
hypertree). Let us turn now to valuations. 
Let $BEL_i$ be the valuation originally attached to the hyperedge $h_i$. 
then $BEL = BEL_1 \odot ...\odot BEL_n$.    (see Fig.\ref{abbvier} a)).
What conditional belief is to be 
attached to $h_n$ ? First marginalize: $BEL'_n = BEL_1^{\downarrow h_1 \cap 
h_n} \odot \dots \odot BEL_{n-1}^{\downarrow h_{n-1} \cap 
h_n} \odot BEL_n$.  (see Fig.\ref{abbvier} b), c))
Now calculate: $BEL"_n={BEL'}_n ^{|h_n \cap h_{i_n}}$, and 
$BEL"'_n={BEL'}_n  ^{\downarrow h_n \cap h_{i_n}}$. 
Let  $BEL_{*k}= BEL_k\oantidot BEL_k ^{\downarrow h_1 \cap 
h_n}$  for k=1,...,$i_n$-1,$i_n$+1,...,(n-1),   (see Fig.\ref{abbvier} d)) 
and 
let $BEL_{*i_n}= (BEL_{i_n}\oantidot BEL_{i_n} ^{\downarrow h_1 \cap 
h_n}) \odot BEL"'_n$ .
Obviously, $BEL=BEL_{*1} 
\odot 
\dots \odot BEL_{*(n-1)} \odot BEL"_n$   (see Fig.\ref{abbvier} e)). 
Now let us consider a new hypertree only with hyperedges $h_1,\dots 
h_{n-1}$, and 
with valuations equal to those marked with asterisk (*), and repeat the 
process 
until only one hyperedge is left, the now valuation of which is considered 
 as $BEL"_1$. In the process, a new factorization is 
obtained: $BEL=BEL"_1 \odot \dots \odot  BEL"_n$. \\
If now for a hyperedge $h_k$ $card(h_k-h_{i_k})=1$, then we assign $BEL"_k$ 
to 
the node of the belief network corresponding to $h_k-h_{i_k}$. If  for a 
 hyperedge $h_k$ $card(h_k-h_{i_k})>1$, then we split  $BEL"_k$ as follows: 
Let $h_k-h_{i_k}=\{X_{k1},X_{k2},....,X_{km}\}$ and the indices shall 
correspond to the order in the belief network induced by the above 
construction procedure. Then 
$$BEL"_k=BEL ^{\downarrow h_k|h_k \cap h_{i_k}}=  
 \bigodot_{j=1}^{m} BEL ^{\downarrow (h_k \cap h_{i_k}) \cup 
\{X_{k1},...,X_{kj}\} | (h_k \cap h_{i_k}) \cup 
\{X_{k1},...,X_{kj}\}-\{X_{kj}\}}$$
and we assign valuation $BEL ^{\downarrow (h_k \cap h_{i_k}) \cup 
\{X_{k1},...,X_{kj}\} | (h_k \cap h_{i_k}) \cup 
\{X_{k1},...,X_{kj}\}-\{X_{kj}\}}$ to the node corresponding to $X_{kj}$ in 
the network structure. It is easily checked that:
\begin{th} \label{xxxx}
(i) The network obtained by the above construction of its structure and 
valuation from hypertree factorization is a belief network.\\
(ii) This belief network represents exactly the joint belief distribution of 
the hypertree\\
(iii) This belief network induces exactly the original reduced hypertree 
structure
\end{th}
The above theorem implies that any hypergraph suitable for 
propagation must have a compatible 
belief network. Hence seeking for belief network decompositions of joint 
belief  distributions  is  sufficient  for  finding  any  suitable 
factorization.
\Sektion{BELIEF TREES AND HYPERTREES}
Let us consider now a special class of hypertrees: connected hypertrees with 
cardinality of each hyperedge equal 2. It is easy to demonstrate that such 
hypertrees correspond exactly to directed trees. Furthermore, valuated 
hypergraphs of this form correspond to belief networks with directed trees as 
underlying dag structures. Hence we can conclude that any factorization in 
form of  connected hypertrees with 
cardinality of each hyperedge equal 2 may be recovered from data by 
algorithms recovering belief trees from data.%
This does not hold e.g. for poly-trees. 

%
Algorithms recovering general type belief networks from data are still to be 
invented. Major obstacle for such algorithms is the badly defined 
relationship 
 between various types of  representation of uncertainty and the empirical 
data. 
This topic will not be discussed here. Instead we will assume that there 
exists a measure $\delta(BEL_1,BEL_2)$ equal to zero whenever both belief 
distributions $BEL_1,BEL_2$ are identical and being positive otherwise. 
Furthermore, we assume that $\delta$ grows with stronger deviation of both 
distributions without specifying it further. 
The algorithm of Chow/Liu \cite{Chow:68} for recovery of tree structure of a 
probability distribution is well known and has been deeply investigated, so 
we will omit its  description. 
It requires  a 
distance measure DEP(X,Y) between each two variables X,Y rooted in 
empirical data and 
spans a maximum weight spanning unoriented tree between the 
nodes. Then any orientation of the tree is the underlying dag structure where 
valuations are calculated  as conditional probabilities.  
 To accommodate it for general belief trees one needs a proper measure of 
distance between variables. As claimed  earlier in \cite{Acid:91}, 
this distance measure has to fulfill  the  following 
requirement: 
$ \min(DEP(X,Y),DEP(Y,Z))>DEP(X,Z)$ for any X, Y, 
 Z such that there exists 
a directed path between X and Y, and between Y and Z.
 For probabilistic belief networks one of such functions is known 
to be Kullback-Leibler distance: 
$$    DEP0(X,Y)=\sum_{x,y} P(x,y)\cdot \log 
\frac{P(x,y)} {P(x)*P(y)} 
$$
If we have the measure $\delta$ available, 
we can construct the measure DEP as follows: 
By the ternary joint distribution of the variables  $X_1,X_3$ with background 
 $X_3$ we understand the function:\\
$$BEL ^{\downarrow X_1 \times X_2[X_3]} 
=(BEL ^{\downarrow X_1 \times X_3 | X_3} \odot
 BEL ^{\downarrow X_2 \times X_3 | X_3} \odot BEL ^{\downarrow X_3})
 ^{\downarrow X_1 \times X_2}$$
Then we introduce:\\
$$DEP_{BN}(X_1,X_2)=
 \min(\delta(BEL ^{\downarrow X_1}\odot  BEL ^{\downarrow X_2}, BEL 
^{\downarrow X_1 \times X_2}),$$
$$, \min_{X_3;X_3 \in \V-{X_1,X_2}}
 \quad \delta(BEL ^{\downarrow X_1 \times 
X_2[X_3]}, BEL ^{\downarrow X_1 \times X_2})) $$\\
with {\V}  being the set of all variables.
The following theorem is easy to prove:
\begin{th}
$ \min(DEP_{BN}(X,Y),DEP_{BN}(Y,Z))>DEP_{BN}(X,Z)$ for any X, Y, Z such that 
there exists a directed path between X and Y, and between Y and Z.
\end{th}
This suffices to extend the Chow/Liu algorithm to recover general belief tree 
networks from data.
 To demonstrate the validity of this general theorem, its specialization was 
implemented 
for the Dempster-Shafer belief networks. The following $\delta$ function was 
 used: Let    $Bel_1$ be a DS belief function and $Bel_2$ be a DS 
pseudo-belief 
function approximating it. Let 
$$\delta(Bel_2,Bel_1)= \sum_{A; m_1(A)>0} m_1(A) \cdot| \ln 
\frac{Q_1(A)}{Q_2(A)}|$$ 
where the assumption is made that natural logarithm of a non-positive number 
is plus infinity.  $|.|$ is the absolute value operator. The values of 
$\delta$ in variable $Bel_2$ with parameter $Bel_1$ range:
$[0,+\infty)$.
For randomly generated tree-like DS belief distributions, if we were working 
directly with these distributions, as expected, the algorithm yielded perfect 
decomposition into the original tree. For random samples generated from such 
distributions, the structure was recovered properly for reasonable sample 
sizes (200 for up to 8 variables). Recovery of the joint distribution was not 
too perfect, as the space of possible value combination is tremendous and 
probably quite large sample sizes would be necessary. It is worth mentioning, 
that even with some departures  from truly tree structure a distribution 
could be obtained which reasonable approximated the original one.
\Sektion{CONCLUSIONS}
In this paper it has been shown        that valuated hypertrees may be 
represented equivalently by belief networks. This is of special importance 
for propagation of uncertainty within the Shenoy/Shafer 
axiomatic framework for local computations. Namely, the simplest possible 
hypertree factorizations of a belief distribution may be obtained by methods 
recovering belief network structure from data and then the valuation of 
hyperedges may be carried out by local computation from marginals of the 
belief distribution. 
Another contribution of this paper is to extend the Chow/Liu algorithm of 
recovering tree-like belief network structures for probability distributions 
onto general type belief distributions fitting the axiomatic framework of 
Shenoy/Shafer.

\end{document}